\documentclass{article}

\PassOptionsToPackage{numbers, compress}{natbib}

\usepackage[final]{neurips_2020}

\usepackage[utf8]{inputenc} 
\usepackage[T1]{fontenc}    
\usepackage{hyperref}       
\usepackage{url}            
\usepackage{booktabs}       
\usepackage{amsfonts}       
\usepackage{nicefrac}       
\usepackage{microtype}      
\usepackage{graphicx}
\usepackage{subcaption}
\graphicspath{ {./images/} }
\usepackage{color}

\title{Deep Learning Models for Predicting Wildfires from Historical Remote-Sensing Data}

\author{
  Fantine Huot\thanks{This work was conducted at Google.} \hspace{0.001in} \textsuperscript{1, 2} \\
  \texttt{fantine@stanford.edu} \\
  \And
  R. Lily Hu\textsuperscript{1}\\
  \texttt{rlhu@google.com} \\
  \And
  Matthias Ihme\textsuperscript{1, 2}\\
  \And
  Qing Wang\textsuperscript{1} \\
  \And
  John Burge\textsuperscript{1} \\
  \And
  Tianjian Lu\textsuperscript{1} \\
  \And
  Jason Hickey\textsuperscript{1} \\
  \And
  Yi-Fan Chen\textsuperscript{1} \\
  \And
  John Anderson\textsuperscript{1} \\
}

\begin{document}

\maketitle

\begin{center}
\textsuperscript{1}Google Research \\
\textsuperscript{2}Stanford University\\
\end{center}

\vspace{0.2in}

\begin{abstract}
Identifying regions that have high likelihood for wildfires is a key component of land and forestry management and disaster preparedness. We create a data set by aggregating nearly a decade of remote-sensing data and historical fire records to predict wildfires. This prediction problem is framed as three machine learning tasks. Results are compared and analyzed for four different deep learning models to estimate wildfire likelihood. The results demonstrate that deep learning models can successfully identify areas of high fire likelihood using aggregated data about vegetation, weather, and topography with an AUC of 83\%.
\end{abstract}

\section{Introduction}
Wildfires can wreak havoc, threatening lives, homes, communities, and natural and cultural resources. Over the last few decades, the wildland fire management environment has profoundly changed, facing longer fire seasons, bigger fires and more acres burned on average each year. In 2019, 775,000 residences across the United States were flagged as at an ``extreme'' risk of destructive wildfire, amounting to an estimated reconstruction cost value of \$220 billion dollars \cite{corelogicwildfire}. Wildland fires have a significant impact on the global climate, representing 8 billion tons or 10\% of global CO\textsubscript{2} emissions per year \cite{vanderwerfglobalfire}. Furthermore, the health impact due to wildfire aerosols is estimated at 300,000 premature deaths globally per year \cite{kollanusmortality}.

For the above reasons, there is a real need for novel wildfire warning and prediction technologies that enable better fire management, mitigation, and evacuation decisions. In particular, assessing the fire likelihood  -- the probability of wildfire burning in a specific location -- would provide valuable insight for forestry and land management, disaster preparedness, and early-warning systems. Additionally, identifying regions of high wildfire likelihood offer the possibility of performing high fidelity but computationally-intensive fluid dynamics simulations specifically on these areas of interest, in order to evaluate  fire-spread scenarios if a fire were to happen.

The increase of remote sensing data availability, computational resources, and advances in machine learning provide unprecedented opportunities  for data-driven approaches to estimate wildfire likelihood. Wildfire likelihood has been based on fire behavior modeling across simulations by varying feature parameters (e.g. weather, topography) that contribute to the probability of a fire occurring {\cite{wildfire_risk_to_communities}}. Instead of using simulations, we train machine learning models on these features to predict the occurrence of wildfires, using data from historical wildfires. 

Recently, deep learning approaches for large-scale remote sensing data have been adopted for physical sciences, with applications ranging from weather forecasting \cite{agrawal2019machine, rasp2020weatherbench}, flood susceptibility \cite{sidrane2019machine}, or real-time fire segmentation from aerial video \cite{doshi2019firenet}. Previous studies of fire risk assessment using machine learning on remote-sensing data used models such as decision trees to estimate fire severity metrics \cite{parks2018high, parks2018mean} and final burn area \cite{coffield2019machine}. Building on this work, we combine historical fire data with deep learning models to take advantage of the feature learning capabilities of deep neural nets.

In this study, we aggregate nearly a decade of satellite observations, combining historical wildfire, terrain, vegetation, and weather data to train a deep learning model. While previous data sets have collected characteristics of fires \cite{short2017spatial,artes2019global}, our data set also includes local geographic data over the area of the fire. Furthermore, compared to other remote sensing wildfire data \cite{sayad2019predictive}, our data set includes richer weather and topography data that are aggregated across multiple data sources, includes sequential data, and also covers a larger geographic area and a larger number of wildfires. We also include corresponding data for negative samples. 

We frame the wildfire likelihood estimation problem as three machine learning problems: image segmentation on daily fire masks, image segmentation on aggregated fire masks, and sequence segmentation. We implement, train, and analyze four different machine learning models: a convolutional autoencoder, a residual U-Net, a convolutional autoencoder with a convolutional LSTM, and a residual U-Net with a convolutional LSTM. We proceed by first describing our data processing pipeline for aggregating historical remote-sensing data in Section~\ref{sec:data}. Then, we present the different deep learning model architectures in Section \ref{sec:model} and discuss our results in \ref{sec:discussion}. 

\section{Data Collection and Processing}
\label{sec:data}

\begin{figure}
  \centering
  \includegraphics[width=1.0\linewidth,clip,trim={.2cm 3.8cm 0cm 0cm}]{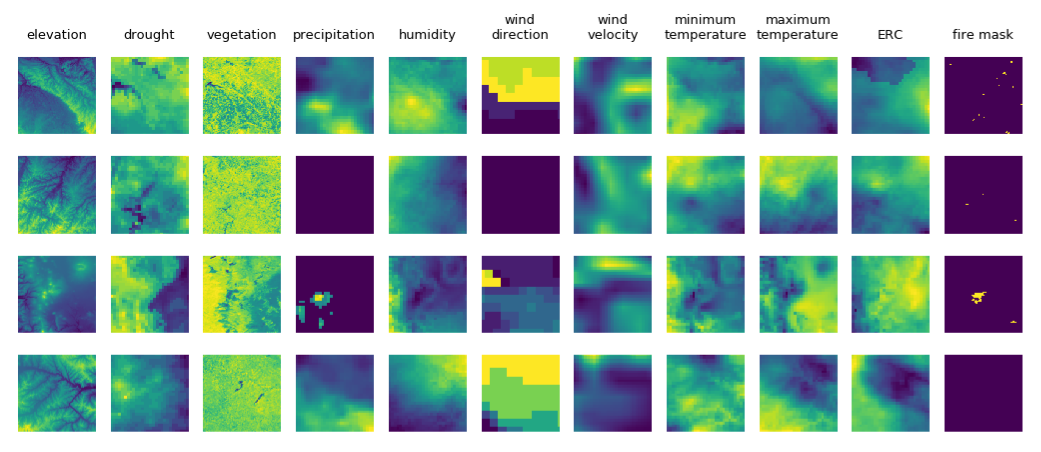}
  \caption{Examples of extracted remote-sensing data from Google Earth Engine. Each row is a sample and each column is a data channel. The last column, ``fire mask'', serves as the label for image segmentation: yellow corresponds to fire, while dark blue corresponds to no fire. Each sample is 128 $\times$ 128 at a resolution of 1km.}
  \label{fig:data_visualization}
\end{figure}

To address the problem of estimating wildfire likelihood, we compile data from multiple remote-sensing data sources from Google Earth Engine (GEE). These data sources are curated for high spatial and temporal resolution, extensive geographical and historical coverage, and to reduce missing data. All the selected data sources are publicly available on Earth Engine, and thus accessible for research and disaster management. These data sources are also regularly updated on Earth Engine with recent data. Logistically, these multiple data sources are already stored on Earth Engine, which allows consistent access across data sources and aligns data sources in location and time.

A use case for models trained on this compiled data is to predict the likelihood of fires given recent remote sensing data. Thus, it is desirable for the data sources to have consistently available data. We selected the following data sources for inclusion:

\begin{itemize}
\item \textbf{Historical wildfire} data are from the MOD14A1 V6 dataset, a collection of daily fire mask composites at 1$\,$km resolution since 2000, provided by NASA LP DAAC at the USGS EROS Center \cite{modis}.
    
\item \textbf{Topography} data are from the Shuttle Radar Topography Mission (SRTM), sampled at 30$\,$m resolution \cite{srtm}.
        
\item \textbf{Drought} is from GRIDMET Drought, a collection of drought indices derived from the GRIDMET dataset, sampled at 4$\,$km resolution every 5 days since 1979, provided by the University of California Merced~\cite{abatzoglou2014seasonal}.
    
\item \textbf{Vegetation} data are from the Suomi National Polar-Orbiting Partnership (S-NPP) NASA Visible Infrared Imaging Radiometer Suite (VIIRS) Vegetation Indices (VNP13A1) dataset, a collection of vegetation indices sampled at 500$\,$m resolution every 8 days since 2012, provided by NASA LP DAAC at the USGS EROS Center \cite{viirs}.
    
\item \textbf{Weather} data are from the University of Idaho Gridded Surface Meteorological Dataset (GRIDMET), a collection of daily surface fields of temperature, precipitation, winds, and humidity at 4$\,$km resolution since 1979, provided by the University of California Merced~\cite{abatzoglou2013development}.
\end{itemize}

These data sources are aggregated and 10 data channels are extracted for their relation to wildfires: elevation, drought index, normalized difference vegetation index (NDVI), precipitation, humidity, wind direction, wind velocity, minimum temperature, maximum temperature, and energy release component (ERC) (see Figure~\ref{fig:data_visualization} for example). All these data channels are available from the data sources cited; none were calculated by us. The data is resampled to 1$\,$km resolution, the resolution of the labels. From this data we sample 128 $\times$ 128 tiles at 1km resolution. The region of study is restricted to the contiguous United States from 2012 to 2020. This time period is selected due to data availability. 

For machine learning, we include all tiles with active fires, where fire pixels that are separated by more than 10$\,$km are considered to belong to a different fire. We also sample twice as many tiles without fire. We split the data between training, evaluating, and testing, by randomly separating all the weeks between 2012 and 2020 according to a 6 : 1 : 1 ratio respectively, while keeping a one-day buffer between weeks from which we do not sample data. The fire masks are treated as segmentation labels of fire versus non-fire, with an additional class for uncertain labels (i.e. cloud coverage or other unprocessed pixels) that will be ignored in the training and evaluation. 

\section{Machine Learning Models} \label{sec:model}

\begin{figure}
\centering
\begin{subfigure}{.25\linewidth}
  \centering
  \includegraphics[width=\linewidth]{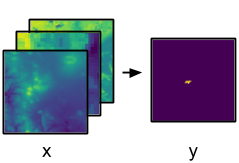}
  \caption{Image segmentation}
  \label{fig:non_aggregated_segmentation}
\end{subfigure}
\hspace{15pt}
\begin{subfigure}{.25\linewidth}
  \centering
  \includegraphics[width=\linewidth]{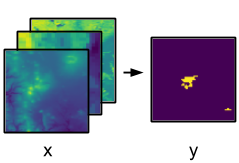}
  \caption{Image segmentation with aggregated labels. }
  \label{fig:aggregated_segmentation}
\end{subfigure}
\hspace{15pt}
\begin{subfigure}{.4\linewidth}
  \centering
  \includegraphics[width=\linewidth]{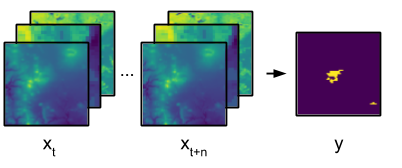}
  \caption{Sequence segmentation}
  \label{fig:sequence_segmentation}
\end{subfigure}%
\caption{The fire likelihood estimation problem is framed as three machine learning tasks: (a) an image segmentation problem, where the input features are taken one day prior to a fire. (b) an image segmentation problem with aggregated fire masks, where the input features are taken one day prior to a fire, and the fire masks are aggregated over a week to capture the total burn area, and (c) a sequence segmentation problem. The input features are arranged in week-long sequences to capture the time-component of the data.}
\label{fig:ml_problems}
\end{figure}

\subsection{Problem Set-Up}
We frame this wildfire prediction problem as three machine learning tasks as illustrated in Figure \ref{fig:ml_problems}. In all the approaches, the remote sensing data is treated as a multi-channel input image and the fire labels is a segmentation map. 

For each of the three tasks, we extract the corresponding datasets:
\begin{enumerate}
\item First, we frame this problem as an image segmentation task. We extract the features one day prior to the corresponding fire masks (Figure \ref{fig:non_aggregated_segmentation}).
\item Since we are interested in the fire likelihood, a snapshot of the fire at a given day does not necessarily provide a good indicator of the actual fire likelihood of the area. Therefore, for the second experiment, we aggregate the fire masks over a week to generate a mask of ``total burn area'' as label for the image segmentation (Figure \ref{fig:aggregated_segmentation}). 
\item Then, to capture the time component of the data, we set up a third experiment to extract the data as week-long daily feature sequences leading to the aggregated fire label (Figure \ref{fig:sequence_segmentation}).
\end{enumerate}

\begin{figure}
\centering
\begin{subfigure}{.23\linewidth}
  \centering
  \includegraphics[clip,trim={1cm 1cm 1cm 1.5cm}, scale=0.14]{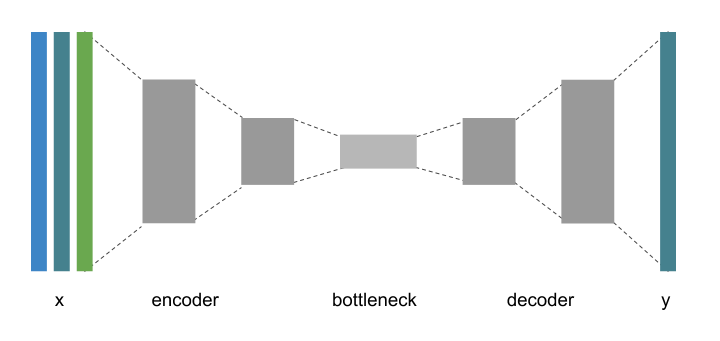}
  \caption{}
  \label{fig:autoencoder}
\end{subfigure}%
\hspace{5pt}
\begin{subfigure}{.23\linewidth}
  \centering
  \includegraphics[clip,trim={1cm 0cm 1cm 1.5cm}, scale=0.14]{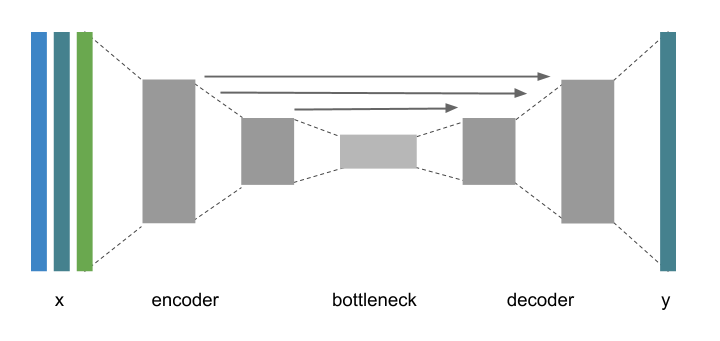}
  \caption{}
  \label{fig:unet}
\end{subfigure}
\begin{subfigure}{.23\linewidth}
  \centering
  \includegraphics[clip,trim={.9cm .5cm 1cm .1cm}, scale=0.3]{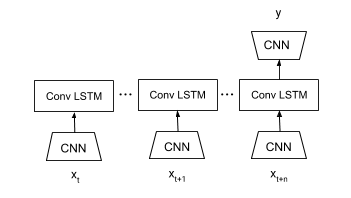}
  \caption{}
  \label{fig:autoencoder_lstm}
\end{subfigure}%
\hspace{5pt}
\begin{subfigure}{.23\linewidth}
  \centering
  \includegraphics[clip,trim={.9cm .5cm 1cm .1cm}, scale=0.3]{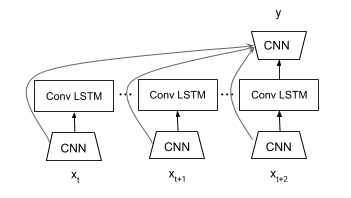}
  \caption{}
  \label{fig:unet_lstm}
\end{subfigure}
\caption{Four different machine learning models for estimating fire likelihood: (a) a convolutional autoencoder, (b) a residual U-Net, (c) a convolutional autoencoder with convolutional LSTM, and (d) a residual U-Net with convolutional LSTM.}
\label{fig:ml_models}
\end{figure}
As a result, the dataset for experiment 1 has about 110,000 samples in the training set, 10,000 samples in the validation set, and 10,000 samples in the testing set. The data sets for experiments 2 and 3 have about 35,000 samples in the training set, 5,000 in the validation set, and 5,000 in the testing set. 

\subsection{Model Architectures}
We implement four machine learning models, as shown in Figure \ref{fig:ml_models}. The detailed architecture for each of these models is provided in the appendix.
\begin{enumerate}
\item A convolutional autoencoder for image segmentation, encoding the features into a bottleneck feature representation before decoding them through upsampling (Figure \ref{fig:autoencoder}). This is a standard approach when both the inputs and outputs are images.
\item A residual U-Net model for image segmentation, similar to the convolutional autoencoder, but with additional skip connections (Figure \ref{fig:unet}). U-Net \cite{ronneberger2015u} is a standard approach for image segmentation, which we augment with short-skip residual connections inspired by ResNet \cite{he2016deep}.
\item A convolutional autoencoder with convolutional LSTM for sequence image segmentation, taking time sequences as inputs (Figure \ref{fig:autoencoder_lstm}). Convolutional LSTMs \cite{shi2015convolutional} are a standard approach for problems with input image sequences.
\item A residual U-Net with convolutional LSTM for sequence image segmentation, taking time sequences as inputs (Figure \ref{fig:unet_lstm}). U-Net improves upon autoencoders for image segmentation, so we replaced the encoder and decoder of model \#3 with the encoder and decoder of our U-Net variant model \#2. 
\end{enumerate}

\subsection{Training Details}
Since wildfires represent only a small area of the total image pixels, we use a weighted cross-entropy loss to overcome the class imbalance. Uncertain labels are ignored in the loss and performance calculations. The models are trained over 100 epochs, with Adam optimizer, on TPUs, using TensorFlow's TPU Distribute strategy. The TPUs we use are Google TPU v3, in an 8-core or 2 $\times$ 2 Pod architecture. Initial experiments were run on both TPUs and V100 GPUs for comparison, but since the results were similar, we use TPUs for the rest of the study. 

We perform the hyperparameter selection by grid search. For each model, we implement the number of layers and the number of filters in each layer as hyperparameters. We explore the number of filters in the residual blocks in the encoder portion of the network according to three different schemes: [64, 128, 256, 256], [32, 64, 128, 256, 256], and [64, 128, 256, 512, 512]. The number of filters in the decoder portion are defined symmetrically. The resulting network architectures are detailed in the appendix. After exploring batch sizes 32, 64, 128, 256, and learning rates 0.01, 0.001, 0.0001, we selected a batch size of 64 and a learning rate of 0.0001. Due to unbalanced classes, weighing the positive labels is essential for the model to estimate fires. After exploring weights of 1, 1.5, 2, 3, 5, and 10, the best results are achieved with a weight of 3. The best model is selected as the one with best AUC (area under the receiver operator curve) on the validation data.

\section{Discussion} 
\label{sec:discussion}

\begin{table}
  \caption{Metrics on the test data for image segmentation with daily active fire labels, aggregated fire labels, and aggregated fire labels with sequential input. Precision, recall, and IoU (Intersection of Union) are for the positive (fire) class.}
  \label{table:results}
  \centering
  \begin{tabular}{lllllll}
    \toprule
    Experiment & Model  & AUC & Precision & Recall & IoU \\
    \midrule
    Daily Segmentation & Autoencoder &  \textbf{0.83} & 0.53 & 0.12 & 0.52\\
    & U-Net & 0.72 & 0.29 & 0.08 & 0.52 \\
    \midrule
    Aggregated Segmentation & Autoencoder &  0.55 & 0.30 & 0.05 & 0.50\\
    & U-Net & 0.54 & 0.29 & 0.05 & 0.50 \\
    \midrule
    Aggregated Segmentation & Autoencoder LSTM &  0.58 & 0.49 & 0.05 & 0.51 \\
    with Sequential input & U-Net LSTM &  0.60 & 0.48 & 0.03 & 0.50 \\
    \bottomrule
  \end{tabular}
\end{table}

\begin{figure}
\centering
\begin{subfigure}{\linewidth}
    \centering
   \includegraphics[clip,trim={0cm 0cm 0cm 0cm}, scale=.48]{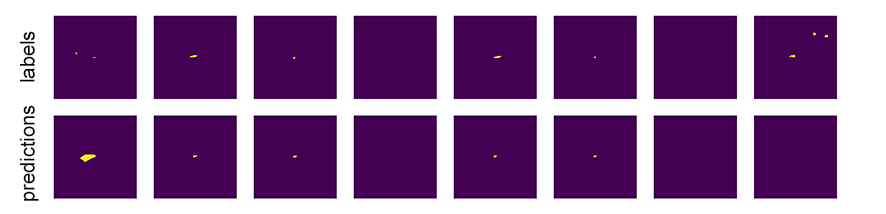}
  \caption{Image segmentation results from the autoencoder model on the test data with daily active fire labels.}
  \label{fig:segmentation_daily}
\end{subfigure}
\begin{subfigure}{\linewidth}
  \centering
   \includegraphics[clip,trim={0cm 0cm 0cm 0cm}, scale=.48]{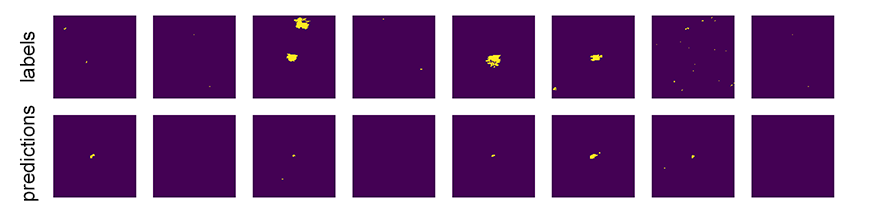}
  \caption{Image segmentation results from the autoencoder LSTM model on the test data with aggregated fire labels.}
  \label{fig:segmentation_aggregated}
\end{subfigure}
\caption{Fire likelihood estimation segmentation results on the test data. The trained models successfully distinguish between tiles with active fires and tiles without fires, and the segmentation errors result from missing small fires, and misclassifications at the boundary between fire and non-fire.}
\label{fig:segmentation}
\end{figure}

The metrics for each of the three problem framings and associated models are presented in Table \ref{table:results}. We found that accuracy across all pixels is not a  useful metric because of the large class imbalance in favor of the negative (no-fire) class. Thus, we compare the metrics on the positive (fire) class. While the metrics on the positive class (here, precision and recall) seem low, the segmentation results in Figure \ref{fig:segmentation_daily} show that fires are detected. The low values are indicative of misclassified pixels at the segmentation boundary between fire and non-fire, or very small fires not being detected, and thus not representative of the overall performance of the models to predict the presence of fire.

The segmentation results with daily fire labels (Figure \ref{fig:segmentation_daily}) demonstrate that deep learning models show real potential for estimating the fire likelihood. The network successfully distinguishes between tiles with active fires and tiles without fires, and the segmentation errors result from missing small fires, and misclassified pixels at the boundary between fire and non-fire. Further experiments can explore smoothing the borders, adding uncertainty borders, and relabeling the fire labels to account for uncertainty in labels.


The segmentation results with aggregated labels (Figure \ref{fig:segmentation_aggregated} and Table \ref{table:results}) are comparably promising. Even when the model does not predict the size of the fire correctly (columns 3 and 5 in Figure \ref{fig:segmentation_aggregated}), it does accurately predict the presence of fire in the given tiles. We expected the aggregated segmentation task would have better performance than the daily segmentation task, but discovered that the opposite occurred. We observed more severe overfitting with the aggregated segmentation map, likely due to the a smaller dataset size. Because the aggregated segmentation maps are created by aggregating the daily segmentation maps over seven days, the resulting aggregated dataset is seven times smaller than the daily dataset. This smaller dataset is easier to overfit on and thus leads to poorer generalization performance. While the experiments with aggregated labels suffer from overfitting and not perform as well overall, the LSTM experiments show that accounting for the time-component does improve the results. These experiments would likely improve from data augmentation techniques, for example, by adding noise to the input features. Another idea would be to increase the time span of the data sets, but this would involve combining multiple data sources for some of the input variables due to limited temporal coverage -- many satellites started collecting data only within the past few years. Another future step is to expand these experiments beyond the United States to a global scale.

Overall, these segmentation results highlight some of the challenges of the fire likelihood estimation task. The image segmentation presents a severe class imbalance because of the small number of fire pixels, even in the tiles with active fires. While the deep learning models can successfully identify fire conditions within a given tile, the actual location remains challenging to predict reliably. To some extent, this behavior is expected when using data from historical fires. Positive events are only labeled when there are actual fires, but the dataset also contains tiles with high fire likelihood that did not have actual fires, such as days before fires or neighboring areas that did not catch fire because firefighters intervened. There has to be some fire precursor for a wildfire to start, which is not necessarily captured by this dataset, especially if the precursor is anthropogenic.

For fire management in practice, the most valuable aspect of this study is deep learning models' ability to estimate which areas are at high risk of large wildfire events. Even without exact segmentation predictions, identifying these regions ahead of time would allow forestry management to allocate resources to specific target regions.  

Expanding this study to more massive datasets would likely improve the prediction results. It could be expanded to a global scale or cover a more extensive time period. Data from recent years, with more fires and larger burnt areas, could be weighted more than data from a decade ago. We could also complement the study with synthetically generated data from high fidelity fire simulations. Additionally, reframing the machine learning problem as a large fire risk assessment task would mitigate errors tied to precise segmentation.  

\section{Conclusions}
We demonstrate the potential of deep learning approaches for estimating the fire likelihood from remote-sensing data. We create a data set by aggregating nearly a decade of remote-sensing data, combining features including weather, drought, topography, and vegetation, with historical fire records. Our trained models can successfully distinguish between fire and non-fire conditions.  Going forward, this data-driven approach could be valuable for wildfire risk estimation, and could be incorporated into wildfire warning and prediction technologies to enable better fire management, mitigation, and evacuation decisions. Beyond the wildfire likelihood problem, the described workflow and methodology could be expanded to other problems such as estimating the likelihood of regions to droughts, hurricanes, and other phenomena from historical remote-sensing data.

\section*{Acknowledgements}
The authors would like to thank Shreya Agrawal, Cenk Gazen, Carla Bromberg, and Zack Ontiveros.

\medskip
\small
\bibliographystyle{unsrtnat}
\bibliography{references}

\medskip

\section*{Appendix A: Model Details}

All convolutions use a kernel size of 3x3 except for skip connections and the last convolution before the output layer.

\begin{figure}
\centering
\begin{subfigure}{.49\linewidth}
  \centering
  \includegraphics[scale=0.16]{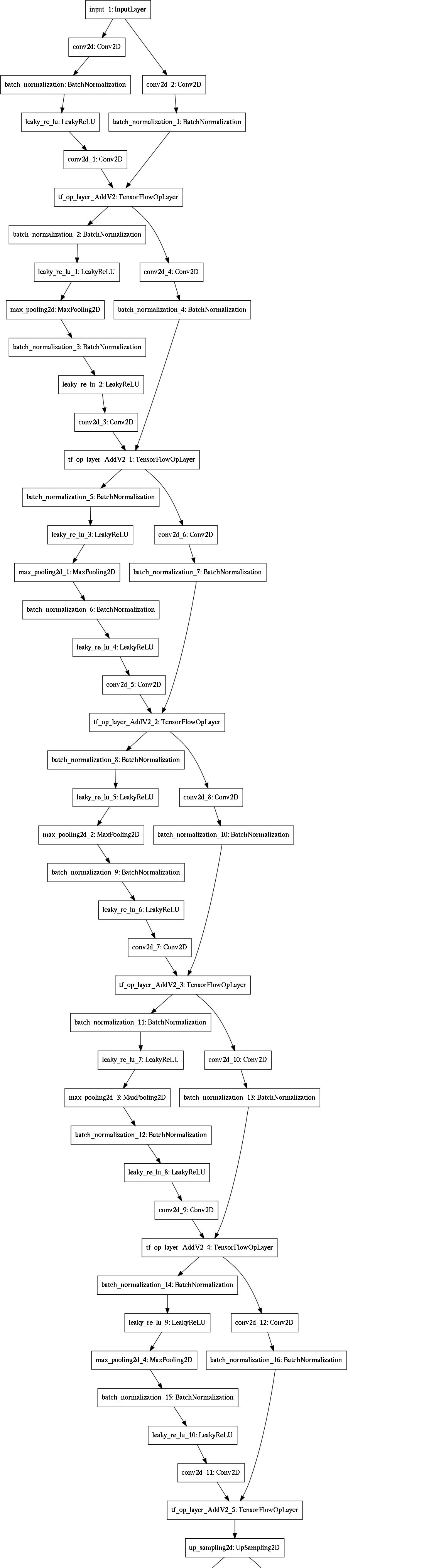}
  \caption{First half of model}
\end{subfigure}
\begin{subfigure}{.49\linewidth}
  \centering
  \includegraphics[scale=0.16]{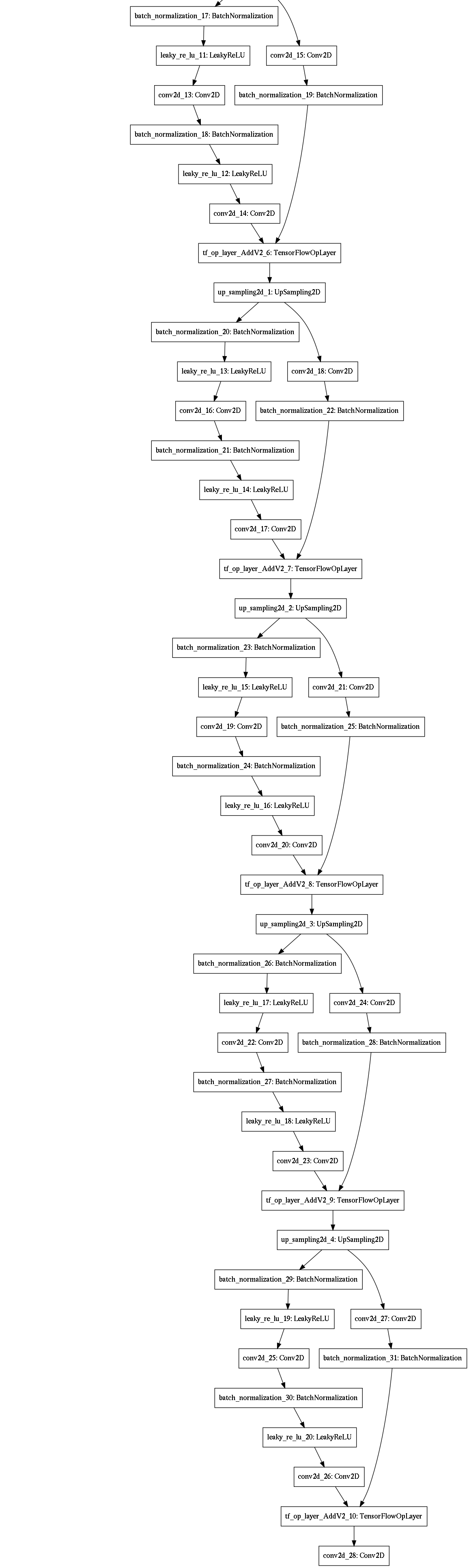}
  \caption*{Second half of model}
\end{subfigure}
\caption*{Figure A.1 Autoencoder model graph}
\label{fig:ml_problems}
\end{figure}

\begin{figure}
\centering
\begin{subfigure}{.49\linewidth}
  \centering
  \includegraphics[scale=0.16]{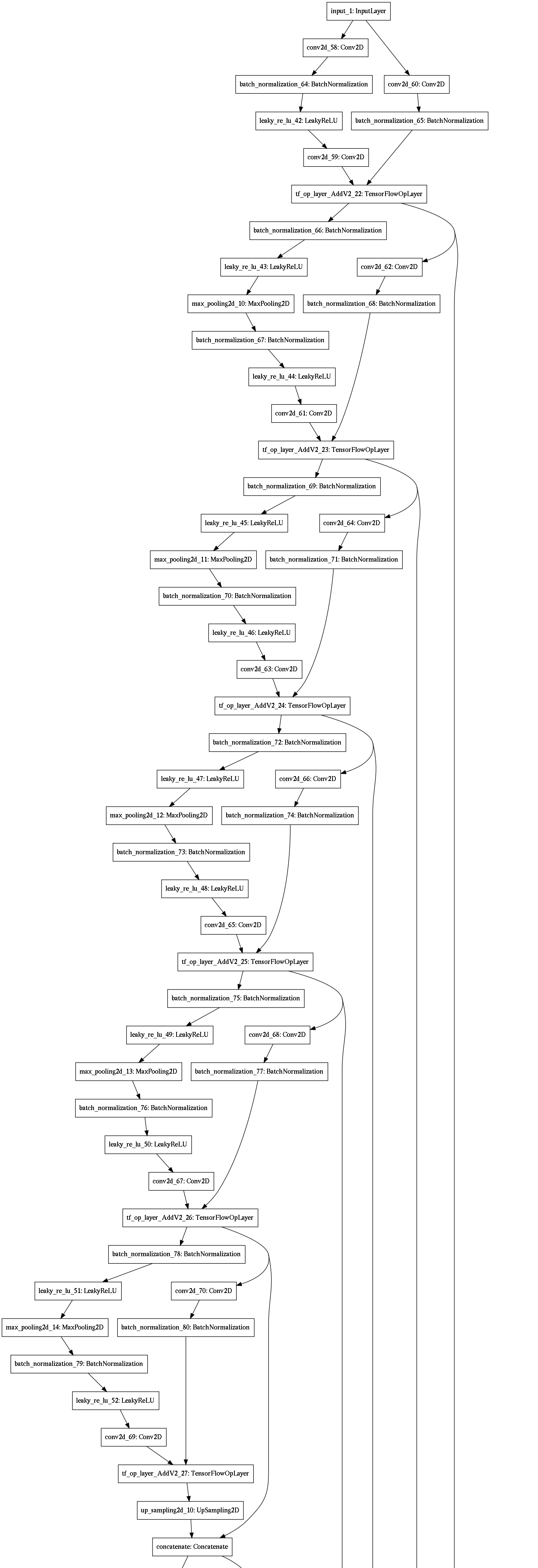}
  \caption{First half of model}
\end{subfigure}
\begin{subfigure}{.49\linewidth}
  \centering
  \includegraphics[scale=0.16]{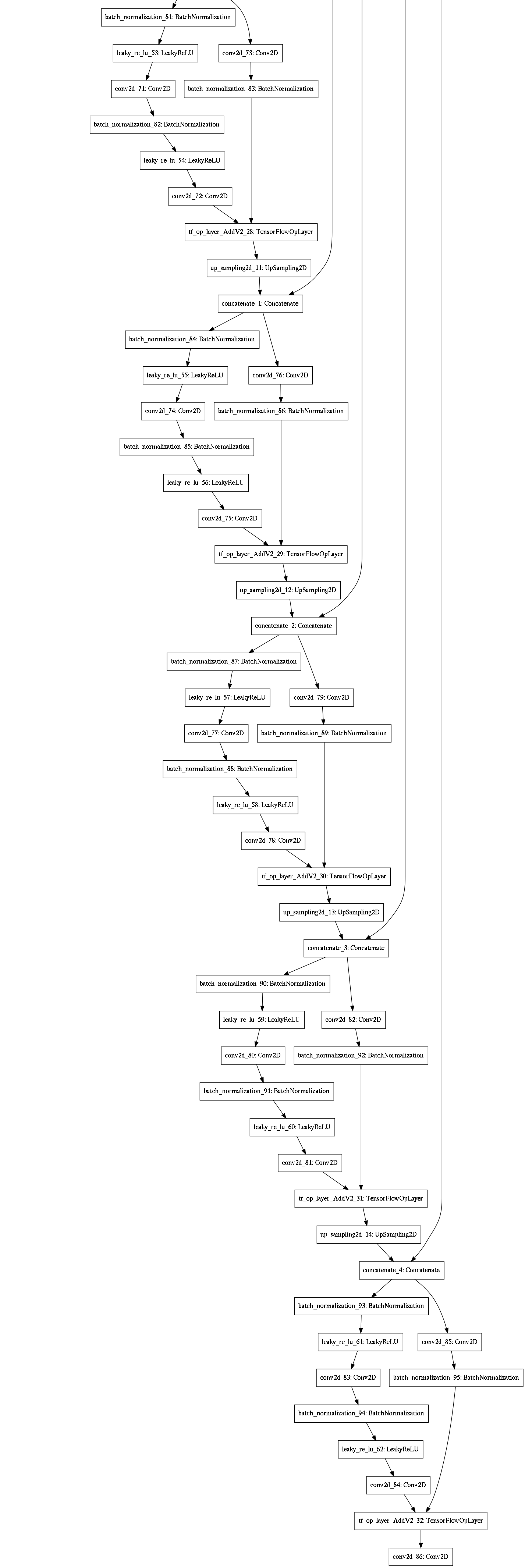}
  \caption{Second half of model}
\end{subfigure}
\caption*{Figure A.2 UNet model graph}
\label{fig:ml_problems}
\end{figure}

\begin{figure}
\centering
\begin{subfigure}{.49\linewidth}
  \centering
  \includegraphics[scale=0.16]{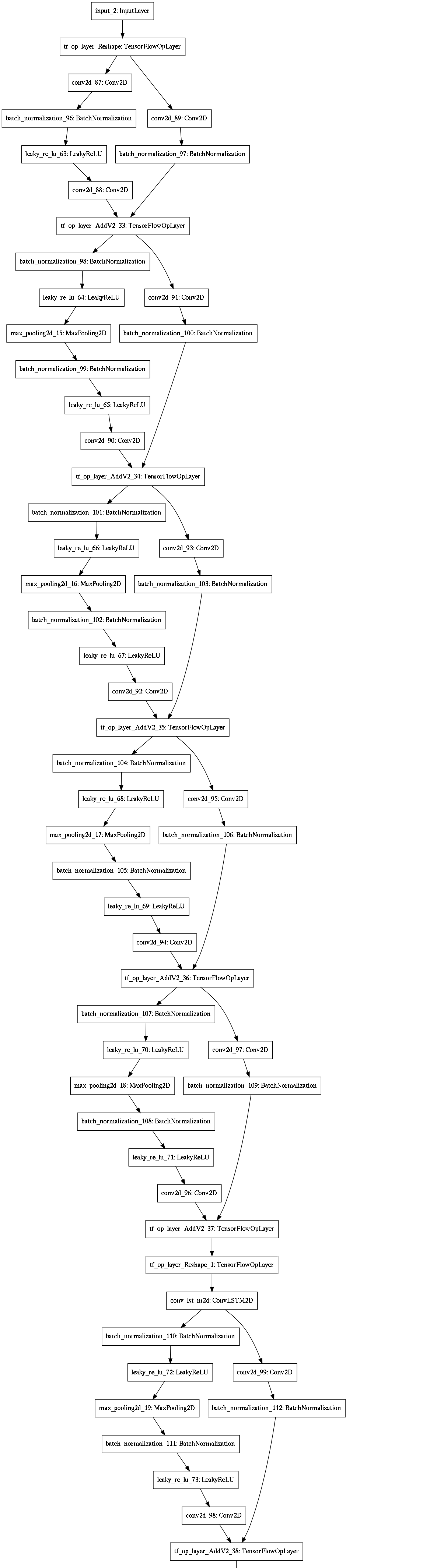}
  \caption{First half of model}
\end{subfigure}
\begin{subfigure}{.49\linewidth}
  \centering
  \includegraphics[scale=0.16]{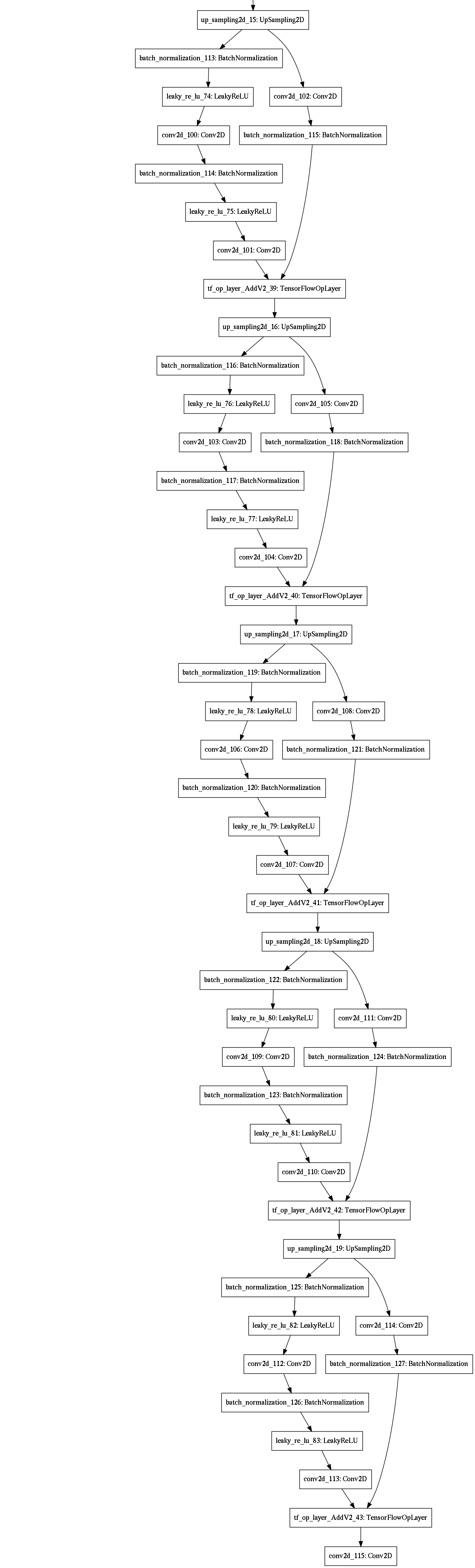}
  \caption{Second half of model}
\end{subfigure}
\caption*{Figure A.3 Autoencoder LSTM model graph}
\label{fig:ml_problems}
\end{figure}

\end{document}